\documentclass[sn-mathphys-num]{sn-jnl}

\usepackage{graphicx}%
\usepackage{multirow}%
\usepackage{amsmath,amssymb,amsfonts}%
\usepackage{amsthm}%
\usepackage{mathrsfs}%
\usepackage[title]{appendix}%
\usepackage{xcolor}%
\usepackage{textcomp}%
\usepackage{manyfoot}%
\usepackage{booktabs}%
\usepackage{array}%
\usepackage{algorithm}%
\usepackage{algorithmicx}%
\usepackage{algpseudocode}%
\usepackage{listings}%
\makeatletter
\providecommand{\headerps@out}[1]{}
\newcommand{\inlinetablecaption}[2]{%
  \refstepcounter{table}%
  \@tablecaption{\fnum@table}{#2}%
  \label{#1}%
}
\makeatother
\providecommand{\footnoteA}[1]{\footnote{#1}}
\newcommand{\analysis}{\textless analysis\textgreater}
\newcommand{\response}{\textless response\textgreater}

\theoremstyle{thmstyleone}%
\theoremstyle{thmstyletwo}%
\theoremstyle{thmstylethree}%

\begin{document}

\title[A Heterogeneous Temporal Memory Governance Framework for Long-Term LLM Persona Consistency]{A Heterogeneous Temporal Memory Governance Framework for Long-Term LLM Persona Consistency}

\author[2]{\fnm{Zhao} \sur{Yang}}\email{zyxiaoqi7@gmail.com}
\equalcont{Zhao Yang and Wang Huan contributed equally as co-first authors.}

\author*[1]{\fnm{Wang} \sur{Huan}}\email{numknw@163.com}

\author[2]{\fnm{Li} \sur{Yingshuo}}\email{lys20020624@gmail.com}

\author[1]{\fnm{Tu} \sur{Haomiao}}\email{THMdeMAILBOX@163.com}

\author[1]{\fnm{Lin} \sur{Hujite}}\email{Felix12369@163.com}

\affil[1]{\orgdiv{School of Electronic Information}, \orgname{Zhongshan Institute, University of Electronic Science and Technology of China}, \orgaddress{\street{No. 1 Xueyuan Road}, \city{Zhongshan}, \postcode{528402}, \state{Guangdong}, \country{China}}}
\affil[2]{\orgname{Changchun Kelaile Technology Co., Ltd}, \orgaddress{\street{Qinglong Road, Luyuan District}, \city{Changchun}, \postcode{130000}, \state{Jilin}, \country{China}}}

\miscnote{Corresponding author: Wang Huan; Zhao Yang and Wang Huan contributed equally as co-first authors.}

\abstract{Large language models (LLMs) often suffer from fact loss, timeline confusion, persona continuity drift, and reduced stability during long-range interactions, especially under high-noise knowledge bases and cross-model transfer settings. To address these issues, we introduce ARPM (Analysis-Based Role-Playing with Memory), an external temporal memory governance framework for long-term dialogue. ARPM physically separates static knowledge memory from dynamic dialogue experience memory. The framework employs a multi-stage pipeline: a retrieval layer that combines vector retrieval, BM25, and RRF fusion; a ranking layer that introduces two temporal coordinates, physical time and dialogue round; and a reading layer that preserves the original semantics and source information of retrieved evidence while organizing candidate content chronologically. This allows the model to access the date of the most recent round and relative round cues. At the generation layer, a controlled <analysis> protocol is used to verify, rerank, and bind candidate evidence to the answer. Unlike approaches that encode persona consistency directly into model weights, ARPM treats long-term continuity as a traceable, auditable, and transferable external governance problem. Using complete engineering logs, we construct three types of experiments. First, under the same 50-round structured question-answering setting, we compare two signal-to-noise ratio conditions, 1:5 and 1:200+, while distinguishing between raw CSV auto-judgment and manual review criteria. Under the 1:5 condition, the original CSV rolling recall accuracy is 54.0\%, whereas manual review raises it to 100.0\%. Under the 1:200+ condition, the corresponding values are 44.0\% and 80.0\%, respectively. These results indicate that automatic rules substantially underestimate truly effective recall when evidence has entered the Prompt and is correctly used by the model. Second, the ablation study shows that dialogue history retrieval is necessary for recent continuity: disabling it reduces strict accuracy from 100\% to 66.7\%. Disabling BM25 reduces strict accuracy to 80.0\%, indicating that pure semantic retrieval is insufficient for long-chain correction and precise tracing. The main role of the dual-temporal mechanism lies in temporal organization and anomaly suppression, rather than in the surface-level correctness of single-turn answers. Third, under a 5.1-million-character noise substrate, periodic context clearing, and multi-model handoff conditions, ARPM maintains relatively high semantic continuity, boundary continuity, and persona consistency across multiple general-purpose model stages, while explicitly exposing boundaries such as insufficient protocol compliance in small models, frame of reference drift, and theatrical drift in dedicated role-playing models. These findings suggest that ARPM’s advantage does not lie in “tuning” a single model into a particular role. Rather, by combining heterogeneous external memory, dual-temporal reranking, temporal evidence unfolding, and analysis-driven evidence verification, ARPM provides a set of transferable continuity conditions for different foundation models. This study demonstrates that long-term persona consistency can be decomposed into engineering components and evaluated in a white-box manner, without relying entirely on the parametric memory of a single model or dedicated role fine-tuning. Systematic memory governance offers a feasible path toward transferable and stable interaction with large language models.}

\keywords{large language models; external memory governance; dual-temporal reranking; high-noise retrieval; persona consistency; cross-model transfer}

\maketitle

\section{Introduction}\label{sec-introduction}

As applications of large language models (LLMs) move from single-turn instruction following and question answering toward long-term agents, personalized assistants, and social chatbot systems, the primary system bottleneck is no longer limited to the parameter scale of the foundation model. It increasingly lies in whether a stable, transferable, and precisely retrievable external memory layer can be built around user interactions, project backgrounds, and dynamic experiences. The technical foundations of modern LLMs are commonly traced to the Transformer architecture, whose self-attention mechanism substantially improved sequence modeling, contextual representation, and large-scale parallel training capabilities \cite{ref11}. Yet the ability to accept longer contexts does not necessarily imply stable use of key evidence within those contexts. Prior work has shown that the position of relevant information in long contexts can substantially affect model reading performance; when evidence appears in the middle of the context, performance may decline markedly \cite{ref16}. This issue is especially salient in long-term companionship, personalized assistant, and complex collaboration scenarios, where consistency in facts, time, identity, and boundaries is challenged simultaneously: the foundation model may be replaced, the context window may be periodically cleared, the external knowledge base may contain large amounts of noise, and users may continually revise previously stated facts. Without appropriate external memory governance, the system is prone to old facts overriding new facts, static knowledge interfering with recent experiences, timeline misalignment, and partial semantic hits that still lead to globally inaccurate answers.

Retrieval-augmented generation (RAG) effectively alleviates hallucinations in static knowledge scenarios \cite{ref1} and has developed into a technical framework centered on the coordinated optimization of retrieval, augmentation, and generation \cite{ref23}. Before RAG, open-domain question answering had already established a “retrieval–reading comprehension” paradigm. For example, DrQA combined Wikipedia retrieval with machine reading comprehension to answer open-domain factual questions \cite{ref24}. REALM further introduced a retriever into language model pretraining, enabling explicit document retrieval for external knowledge support \cite{ref12}. DPR showed that dense dual-encoder retrieval can effectively replace traditional sparse retrieval in open-domain question answering and improve candidate passage recall \cite{ref13}. However, these lines of work still typically optimize for whether a single question can be answered correctly, and pay relatively limited attention to two core challenges in long-term continuous interaction. First, memory types are inherently heterogeneous: objective knowledge, chat experiences, user corrections, style boundaries, and task states operate at different temporal scales and reliability levels. Second, evidence use before and after generation can diverge: in real systems, even when supporting evidence has already entered the Prompt, automatic scoring rules may still mark a round as a “retrieval error” because of a top-1 miss, field-matching failure, or overly strict criteria, thereby underestimating the system’s actual capability.

The problem is therefore not merely whether a particular model resembles a certain persona. It is whether, under the simultaneous presence of a noisy knowledge base, context discontinuity, and model replacement, the system can continue the same narrative, the same set of key experiences, and the same interaction boundaries. To address this problem, this paper proposes ARPM. Rather than relying on the parametric memory of a single model for long-term continuity, ARPM decomposes the problem into four governable components: heterogeneous dual-source memory decoupling, dual-temporal ranking, analysis-driven evidence verification, and white-box logging with manual review. In this formulation, the system’s “memory” is no longer a single hit in a vector database, but a complete closed loop from candidate retrieval, evidence entry, and generation constraints to post-hoc auditing.

Building on this methodological formulation, the paper implements the main line of “heterogeneous decoupling + dual-temporal modeling + metacognitive gate” through reproducible engineering logs and quantifiable experimental design. The method is therefore not only a conceptual framework, but also a traceable and testable process for continuity governance. Rather than making the abstract claim that “the system is more human-like,” this study uses real data to answer three specific questions: (1) Can ARPM go beyond mechanical top-1 rules and complete secondary utilization after evidence enters the Prompt? (2) Without semantic compression or graph-based rewriting, does unfolding retrieved evidence in chronological order and explicitly exposing the date of the most recent round help the model establish a more stable frame of reference and improve its analysis process? (3) Can long-term continuity transfer across models with the support of external memory and temporal anchoring, and what clearly observable boundaries remain?

\subsection{Main Contributions}\label{subsec-main-contributions}

The contributions of this work are fourfold:

\begin{enumerate}[1.]
\item A heterogeneous dual-source external memory governance framework. This paper physically separates static knowledge memory from dynamic dialogue experience memory and merges them at the candidate retrieval stage. This structure alleviates retrieval competition among static knowledge, old experiences, and recent facts during long-range interactions.
\item A coordinated mechanism for dual-temporal ranking, temporal evidence unfolding, and hybrid retrieval. By jointly modeling physical time and dialogue round, this paper combines vector retrieval, BM25, and RRF fusion ranking. When candidate evidence enters the Prompt, the framework preserves the original semantics and source information, organizes retrieved content chronologically, and explicitly exposes the date of the most recent round and relative round cues. This enables the system to obtain more stable evidence candidates and a clearer frame of reference under high-noise knowledge bases, recent fact recovery, and time-sensitive follow-up questions.
\item An analysis-driven evidence verification mechanism. This paper defines the \analysis{} protocol as a controlled verification and binding mechanism after retrieval and before generation, rather than as an open-ended long Chain-of-Thought. This mechanism helps explain why, in some rounds, automatic rules fail to hit while the evidence that has entered the Prompt is still correctly used by the model.
\item A cross-model continuity validation process. This paper conducts multi-stage handoff experiments under high-noise, context-clearing, and foundation-model switching conditions. The results show that long-term persona consistency is not fully bound to the weights of a single model, but can achieve relatively strong cross-model transfer through the joint effect of external memory, temporal anchoring, and generation constraints, while explicitly exposing its capability boundaries.
\end{enumerate}

\section{Related Work}\label{sec-related-work}

\subsection{Retrieval-Augmented Generation and Long-Dialogue Retrieval Problems}\label{subsec-retrieval-augmented-generation-and-long-dialogue-retrieval-problems}

Conventional RAG primarily targets static knowledge question answering, with optimization focused on improving external fact retrieval and answer correctness \cite{ref1}. In the broader development of open-domain question answering, early systems such as DrQA combined large-scale document retrieval with reading comprehension, establishing the basic “retrieve first, then read” paradigm \cite{ref24}. REALM introduced retrieval augmentation into language model pretraining, making external knowledge access part of the model’s capability \cite{ref12}. DPR further improved candidate passage recall in open-domain question answering through dense dual-encoder representations \cite{ref13}. In recent years, RAG has evolved from simple retrieval-augmented question answering into more complex system forms, including Advanced RAG and Modular RAG \cite{ref23}.

However, when the task shifts to long-term dialogue, the central challenge is no longer merely whether a relevant document can be retrieved, but which type of memory should be prioritized at the current moment. If static knowledge, recent user experiences, historical task states, and role-boundary texts are mixed into the same index, the system is prone to retrieval competition: older knowledge that appears semantically closer may override a newly corrected user fact. Term-matching methods such as BM25 remain effective for hard fact localization \cite{ref2}, while ranking fusion methods such as RRF can integrate heterogeneous retrieval signals \cite{ref3}. Building on this line of work, this paper further emphasizes that the challenges of long-term memory systems arise not only at the retrieval end, but also propagate to evidence usage bias in the reading phase and generation stage.

\subsection{Memory-Augmented Large Language Models}\label{subsec-memory-augmented-large-language-models}

To address long-context limitations, prior work has proposed various external memory mechanisms, including hierarchical memory management, multi-level paging, event extraction, temporal decay, and dialogue summarization \cite{ref4,ref5,ref6,ref7,ref8}. These methods demonstrate the feasibility of externalizing memory. However, in scenarios involving long-term persona and temporal continuity, a single timeline and a mixed index are often still insufficient. Moreover, long context itself does not necessarily constitute reliable memory: previous studies have shown that language models are sensitive to the position of evidence in long contexts. Relevant information is typically easier to use when it appears at the beginning or end of the context, whereas performance may decline when it appears in the middle \cite{ref16}.

This indicates that long-term memory systems need to address not only indexing and retrieval, but also how candidate evidence is organized, ranked, and read in the Prompt. Under conditions such as user-corrected facts, cross-day dialogue, and periodic context clearing, relying solely on similarity-based retrieval can easily lead to frame of reference misalignment. ARPM differs in that it simultaneously introduces two temporal dimensions—physical time and dialogue round—and uses them for ranking competition and evidence organization in the reading phase, rather than treating them merely as metadata for post-hoc archiving.

\subsection{Persona Consistency and Cross-Model Continuity}\label{subsec-persona-consistency-and-cross-model-continuity}

Research on persona consistency generally follows two paths. One line of work solidifies role style and boundaries into model weights through supervised fine-tuning or preference optimization. The other relies on a system prompt or setting document for continuous control \cite{ref9,ref10}. In open-domain dialogue research, works such as PersonaChat use persona profile-conditioned generation, enabling chatbots to produce more specific and consistent responses around given identity information \cite{ref9}. Profile Consistency Identification further focuses on identifying persona-profile consistency in open-domain dialogue \cite{ref10}. In addition, some studies formulate dialogue consistency as a natural language inference problem and construct the Dialogue NLI dataset to determine whether persona settings and model responses stand in entailment, contradiction, or neutrality relations \cite{ref21}. From the broader perspective of social chatbot research, user engagement, emotional connection, and persona continuity in long-term interaction have long been important goals in dialogue system design \cite{ref25}.

However, methods based on role fine-tuning have limited transferability when the foundation model changes, while methods based on prompts or setting documents tend to degrade as dialogue length increases and context is cleared. In contrast, ARPM addresses the problem of external continuity governance: whether the system can maintain the same fact chain, task chain, and role boundary chain after model replacement. This paper separates “remembering” from “being like the same persona” and argues that long-term persona consistency is not a single memory accuracy score, but the joint outcome of fact retrieval, temporal judgment, protocol compliance, language style, and boundary behavior.

\section{Method}\label{sec-method}

\subsection{Overall Design Objective}\label{subsec-overall-design-objective}

ARPM is not intended to make the model “better at performing” across all scenarios. Instead, it decomposes long-term continuous interaction into an engineering process that is governable, observable, and reproducible. The core system consists of four components: heterogeneous dual-source memory decoupling, dual-temporal ranking and temporal evidence unfolding, analysis-driven evidence verification, and white-box log write-back.

The basic workflow is shown in Figure 1. A user query first undergoes lightweight query augmentation and is then routed separately to the knowledge memory route and the experience memory route. Candidates from the two routes are merged under dual-temporal constraints, and the retrieved content is organized chronologically while preserving its original semantics. After entering the Prompt, the evidence is verified and bound to the answer through a controlled \analysis{} protocol. Finally, the response and intermediate states are written into structured logs.

\begin{figure}[h]
\centering
\includegraphics[width=0.95\textwidth]{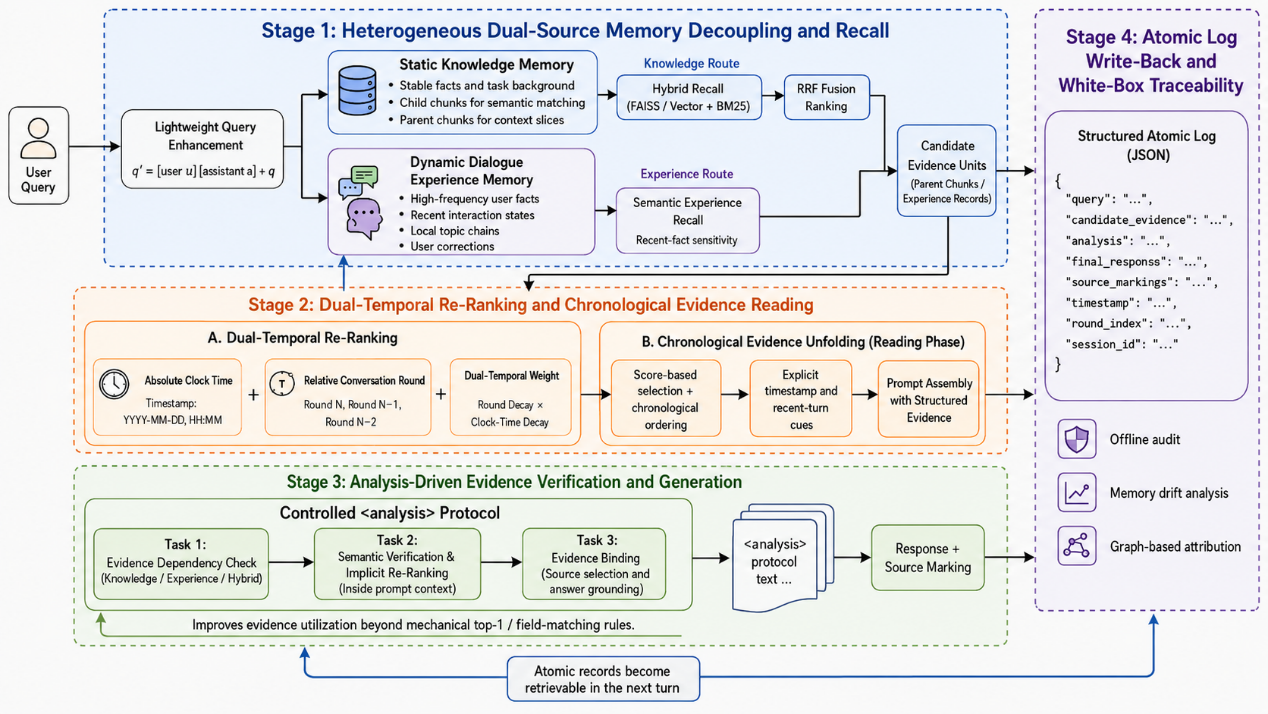}
\caption{Overview of the ARPM method: heterogeneous dual-source retrieval, dual-temporal ranking, temporal evidence unfolding, and the analysis-driven evidence verification pipeline.}\label{fig1}
\end{figure}

\subsection{Heterogeneous Dual-Source Memory Decoupling}\label{subsec-heterogeneous-dual-source-memory-decoupling}

ARPM divides external memory into two types. The first is knowledge memory, which stores relatively stable, low-frequency objective knowledge or task background. The second is experience memory, which stores high-frequency user facts, interaction states, and local topic chains. Knowledge memory adopts a parent-child chunk structure: child chunks are used for fine-grained semantic matching, while parent chunks provide more complete context slices to the generation stage. Experience memory directly supports recent topic continuation and the recovery of user-corrected facts.

This physical separation does not mean that the two routes never interact; rather, it prevents them from mixing prematurely during retrieval. The knowledge route introduces hybrid retrieval with BM25 and vector retrieval \cite{ref2}, followed by RRF fusion ranking to handle high-noise factual question answering \cite{ref3}. The vector retrieval route follows the same basic idea as dense retrieval methods such as DPR: candidate evidence is obtained through nearest-neighbor search in a semantic vector space \cite{ref13}. In the engineering implementation, FAISS provides a mature tool foundation for large-scale vector similarity search, index construction, and approximate nearest-neighbor retrieval \cite{ref15}. The experience route remains sensitive to colloquial short references, local topic chains, and recent fact updates. What ultimately enters the Prompt is not a single top-1 result, but a combination of candidates after reranking in the two routes. To keep the method description from remaining purely conceptual, the main formulas corresponding to the current engineering implementation are given below.

The knowledge route first computes the vector inner-product score:

\begin{equation}
s_{\mathrm{vec}}(q, d) = q \cdot  d\tag{1}\label{eq1}
\end{equation}

It then computes a BM25+-style keyword score:

\begin{equation}
s_{\mathrm{BM25}}(q,d) = \sum_{t\in q} \operatorname{idf}(t) \cdot \left[\frac{\operatorname{tf}(t,d)(k_1+1)}{\operatorname{tf}(t,d)+k_1(1-b+b\cdot |d|/\operatorname{avgdl})}\right]\tag{2}\label{eq2}
\end{equation}

where $\operatorname{idf}(t)=\max\left(0,\log\frac{N-\operatorname{df}(t)+0.5}{\operatorname{df}(t)+0.5}\right)+\delta$. This implementation uses a BM25 variant with positively biased IDF, denoted as a BM25+-style score.

Vector retrieval and BM25 retrieval are fused through Reciprocal Rank Fusion:

\begin{equation}
s_{\mathrm{RRF}}(d)=\sum_i \frac{1}{k_{\mathrm{rrf}}+\operatorname{rank}_i(d)+1}\tag{3}\label{eq3}
\end{equation}

The base score of the knowledge route can be written as:

\begin{equation}
s_{\mathrm{kb}}^0(d)=\operatorname{norm}(s_{\mathrm{RRF}}(d))+b_{\mathrm{user}}+b_{\mathrm{character}}+b_{\mathrm{source}}\tag{4}\label{eq4}
\end{equation}

The dialogue history route uses normalized cosine similarity as the semantic base score:

\begin{equation}
s_{\mathrm{sem}}(q, d) = \operatorname{clip}(\cos(q, d), 0, 1)\tag{5}\label{eq5}
\end{equation}

\begin{equation}
s_{\mathrm{chat}}^0(d)=s_{\mathrm{sem}}(q,d)+b_{\mathrm{session}}+b_{\mathrm{user}}+b_{\mathrm{character}}\tag{6}\label{eq6}
\end{equation}

\subsection{Lightweight Query Augmentation, Dual-Temporal Ranking, and Temporal Evidence Unfolding}\label{subsec-lightweight-query-augmentation-dual-temporal-ranking-and-temporal-evidence-unfolding}

In real logs, ARPM applies lightweight role-prefix enhancement to the query, namely:

\[
q' = [\text{User } u][\text{Assistant } a] + q
\]

This design does not perform the main task of “understanding.” Its purpose is to provide the minimum necessary interaction-role information for the knowledge route and the experience route. The more critical processing occurs in the ranking and reading stages. The system maintains two temporal coordinates for each memory chunk: absolute physical time and relative dialogue round. The former indicates when an event occurred, while the latter indicates how far it is from the current topic.

ARPM performs dual-temporal reranking separately in the two routes. In the knowledge route, retrieval fusion and role weighting are completed first, followed by temporal weighting. In the dialogue history route, semantic matching and role weighting are completed first, followed by temporal weighting.

The round decay term and physical-time decay term are defined as follows:

\begin{equation}
w_{\mathrm{round}}(d)=\exp\left(-\frac{|r_{\mathrm{current}}-r_d|}{\lambda_{\mathrm{round}}}\right)\tag{7}\label{eq7}
\end{equation}

\begin{equation}
w_{\mathrm{clock}}(d)=\exp\left(-\frac{\Delta h_d}{\lambda_{\mathrm{hours}}}\right)\tag{8}\label{eq8}
\end{equation}

The dual-temporal retention weight is the product of the two:

\begin{equation}
w_{\mathrm{temporal}}(d)=w_{\mathrm{round}}(d)\cdot w_{\mathrm{clock}}(d)\tag{9}\label{eq9}
\end{equation}

In the two routes, temporal reranking acts on their respective base scores:

\begin{equation}
s_{\mathrm{kb}}(d)=s_{\mathrm{kb}}^0(d)\cdot w_{\mathrm{temporal}}^{\mathrm{kb}}(d)\tag{10}\label{eq10}
\end{equation}

\begin{equation}
s_{\mathrm{chat}}(d)=s_{\mathrm{chat}}^0(d)\cdot w_{\mathrm{temporal}}^{\mathrm{chat}}(d)\tag{11}\label{eq11}
\end{equation}

In the current engineering implementation, the two routes perform temporal reranking separately. Under the default configuration, the same decay constants are often used, with $\lambda_{\mathrm{round}}$ = 20 and $\lambda_{\mathrm{hours}}$ = 168 hours. However, their weighting and competition processes are independent and can therefore be adjusted separately according to the experimental setting.

\subsection{Analysis-Driven Evidence Verification and Reranking}\label{subsec-analysis-driven-evidence-verification-and-reranking}

We treat the \analysis{} protocol as a controlled pre-generation verification mechanism rather than an open-ended reasoning chain. Prior work on Chain-of-Thought prompting has shown that explicit intermediate reasoning steps can improve the performance of LLMs on complex tasks \cite{ref17}, and self-consistency reasoning further suggests that selecting among multiple candidate reasoning paths based on consistency can reduce incidental errors introduced by any single path \cite{ref18}. However, the \analysis{} protocol in ARPM is specifically constrained as a post-retrieval and pre-generation mechanism for evidence verification, implicit re-ranking, and answer binding. This design is methodologically related to reasoning-and-acting frameworks such as ReAct, where the model does not rely solely on internal parameters, but establishes a more interpretable closed loop among explicit evidence access, local reasoning, and output constraints \cite{ref19}. Consequently, the \analysis{} stage does not replace retrieval. Instead, after candidate evidence has entered the Prompt and has been organized chronologically, it performs three core operations:

Assessing evidence requirements: determining whether the current question relies more on knowledge evidence, experience evidence, or a combination of both;

Semantic verification and implicit re-ranking: performing semantic verification and implicit re-ranking among multiple parent chunks that have entered the Prompt;

Answer binding: binding the final answer to interpretable evidence sources.

The “CoT noise-resistance capability” emphasized by ARPM does not mean allowing the model to reason without grounding. Rather, it aims to reduce erroneous dependencies and improve evidence utilization within the existing candidate context. This design helps explain the discrepancy observed in the high-noise experiment: although the original CSV-based automatic rules frequently marked certain rounds as 0, manual review found that the corresponding supporting parent chunks had already entered the Prompt, and that the model’s answer was semantically consistent with the evidence. This phenomenon directly reflects the joint effect of the \analysis{} protocol and temporal evidence unfolding.

\subsection{Real-Time Atomic Write-Back and White-Box Traceability Mechanism}\label{subsec-real-time-atomic-write-back-and-white-box-traceability-mechanism}

Real-time atomic distillation and the white-box traceability mechanism constitute one of the key engineering components of this paper. In the current implementation, the system does not simply persist raw dialogue as monolithic logs. Instead, for each interaction round, it writes the query, candidate parent chunks, pre-generation analysis, final response, and source markings into logs and indexes as structured atomic units, thereby providing traceable candidates for retrieval in the next round. In other words, ARPM’s external memory is not a long text summarized after the fact, but a set of atomic records with source, temporal, and path markings. These records preserve their original semantics when read, without compression-based rewriting or graph-based transformation.

To ensure that the above heterogeneous decoupling and temporal governance mechanisms can be strictly executed and quantitatively evaluated, ARPM is instantiated at the engineering level as a modular monolithic Web service architecture. The frontend provides a multi-turn dialogue interface and experimental intervention interface, while the backend completes the data closed loop from query augmentation, dual-source retrieval, and dual-temporal ranking to response generation and write-back auditing. To address the difficulty of fine-grained attribution in long-range LLM evaluation, which often relies heavily on black-box logs from commercial APIs, the system writes all key fields to disk in real time using a standardized JSON structure. These fields include chunk\_id, session\_id, user\_input, assistant\_reply, source\_type, physical timestamp, and relative dialogue round. This design is partly consistent with the idea of tool-augmented language models: the model does not rely entirely on internal parametric capabilities, but obtains more reliable task-execution conditions through external tools, retrieval systems, APIs, or structured logging mechanisms \cite{ref22}.

ARPM’s external memory layer therefore supports not only retrieval augmentation, but also evidence tracing, error attribution, and state write-back. This traceability paradigm enables offline replay and graphical attribution, providing a white-box auditing basis for memory drift, changes in temporal weights, and error-path identification. Baseline sampling shows that the system has recorded more than 1,000 atomic memory blocks, covering 6 independent session segments and 9 ablation session segments. The main session depth is approximately 180 rounds, and the number of atomic chunks in a single session is approximately 50–180. In complex long-range contexts, responses containing \analysis{} tags account for approximately 85\%–95\%. These logs are not supplementary materials, but the foundation of the experimental design and the conclusions of this paper.

\section{Experimental Design}\label{sec-experimental-design}

\subsection{Implementation and Logging Baseline}\label{subsec-implementation-and-logging-baseline}

ARPM was configured using parameters drawn from real operational logs. In a representative setting, \textasciigrave{}knowledge\_k\textasciigrave{} was set to 5, \textasciigrave{}chat\_history\_k\textasciigrave{} to 10, \textasciigrave{}similarity\_threshold\textasciigrave{} to 0.5, and \textasciigrave{}RRF\_k\textasciigrave{} to 60, with both BM25 and RRF enabled. Role enhancement and temporal decay were applied independently to the knowledge route and the experience route. Top-K was not fixed across all settings: the high-noise retrieval experiment used a forced 3+3 retrieval configuration, while the other experiments generally used 5+10.

BM25 was retained in the knowledge route to support precise factual localization under high-noise conditions. The experience route, by contrast, was designed to emphasize recent experience retrieval and contextual continuity. For every experiment, we preserved complete pipeline logs, so that manual review could be grounded in the corresponding \textasciigrave{}parent\_id\textasciigrave{}, evidence text, and generated output, rather than in the final answer alone.

\subsection{High-Noise Retrieval Experiment}\label{subsec-high-noise-retrieval-experiment}

The high-noise experiment was conducted on the same 50-round structured question-answering set under two signal-to-noise ratio settings: 1:5 and 1:200+. In this context, the signal-to-noise ratio refers to the relative proportion of valid knowledge to noise in the knowledge base substrate. The 1:200+ condition therefore introduced a much stronger level of interference from irrelevant knowledge.

To avoid treating automatic scoring rules as a proxy for the system’s actual capability, we computed two curves in parallel:

\begin{equation}
\mathrm{CSV\_RollingAccuracy}(r)=\frac{1}{r}\cdot\sum_{i=1}^{r}\mathrm{AutoCorrect}_i\tag{13}\label{eq13}
\end{equation}

\begin{equation}
\mathrm{Manual\_RollingAccuracy}(r)=\frac{1}{r}\cdot\sum_{i=1}^{r}\mathrm{FinalCorrect}_i\tag{14}\label{eq14}
\end{equation}

\begin{equation}
\begin{aligned}\mathrm{FinalCorrect}_i &= 1 \text{ if and only if } \mathrm{ManualSupport}_i = 1 \text{ and } \mathrm{AnswerCorrect}_i = 1;\\\mathrm{FinalCorrect}_i &= 0 \text{ otherwise.}\end{aligned}\tag{15}\label{eq15}
\end{equation}

The automatic curve captures whether the scoring rule directly identifies the supporting evidence as valid. The manual curve instead asks whether the evidence included in the Prompt can in fact support the model’s answer. This distinction is important for identifying analysis-driven evidence re-utilization, and for avoiding the conflation of mechanical top-1 matching with the system’s actual evidence-reading capability.

\begin{center}
\inlinetablecaption{tab1}{Cumulative recall results under automatic judgment and manual review across two signal-to-noise ratio settings}
\centering
\scriptsize
\setlength{\tabcolsep}{3pt}
\renewcommand{\arraystretch}{1.25}
\begin{tabular}{|>{\raggedright\arraybackslash}p{0.1012\textwidth}|>{\raggedleft\arraybackslash}p{0.0916\textwidth}|>{\raggedleft\arraybackslash}p{0.1096\textwidth}|>{\raggedleft\arraybackslash}p{0.1330\textwidth}|>{\raggedleft\arraybackslash}p{0.0861\textwidth}|>{\raggedleft\arraybackslash}p{0.1170\textwidth}|>{\raggedleft\arraybackslash}p{0.1018\textwidth}|>{\raggedleft\arraybackslash}p{0.1096\textwidth}|}
\hline
Signal-to-noise ratio setting & Test rounds & Correct rounds under original CSV & Original CSV rolling accuracy & Correct rounds under manual review & Manual rolling accuracy & 0-to-1 correction rounds & Rounds pending further review\\
\hline
1:5 & 50 & 27 & 54.0\% & 50 & 100.0\% & 23 & 0\\
\hline
1:200+ & 50 & 22 & 44.0\% & 40 & 80.0\% & 18 & 10\\
\hline
\end{tabular}
\end{center}

\begin{figure}[h]
\centering
\includegraphics[width=0.95\textwidth]{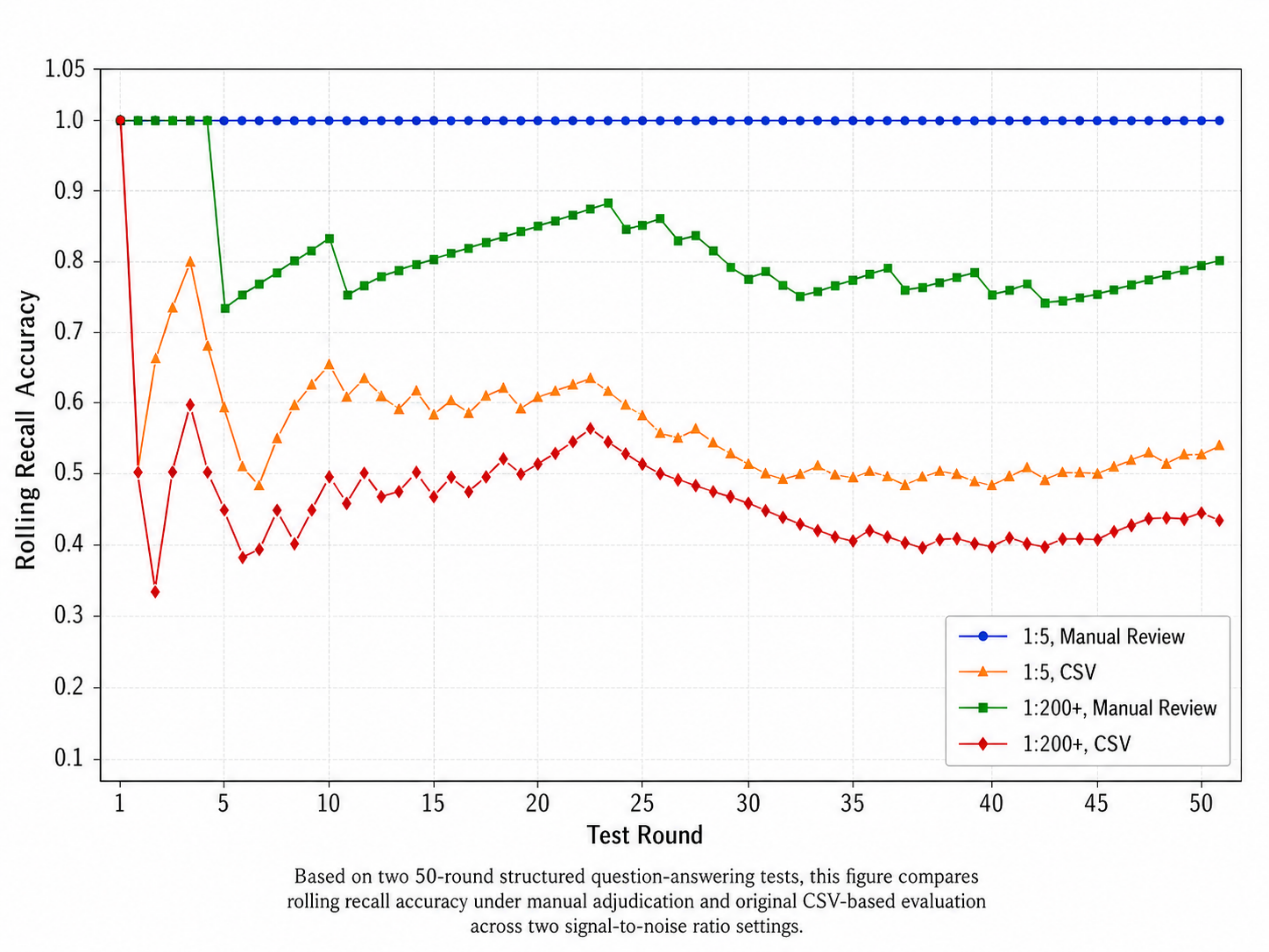}
\caption{Comparison of rolling recall accuracy under two signal-to-noise ratio settings: the manual review curve is substantially higher than the original CSV-based automatic curve.}\label{fig2}
\end{figure}

\subsection{Ablation Study}\label{subsec-ablation-study}

To evaluate the contribution of individual components, we conducted an ablation study using the full system as the baseline. The analysis compared five key components: knowledge base retrieval, dialogue history retrieval, the dual-temporal ranking mechanism, hybrid retrieval, and automatic repair and protocol tightening. Pure dialogue mode and the strong consistency preset were also included as control conditions. Performance was assessed using two primary metrics: strict accuracy, which measures whether a question can be correctly recovered under the given context and mechanism constraints, and anomaly rate, which captures failures such as temporal-priority confusion, loss of correction nodes, or breaks in the character or task chain.

\begin{center}
\begin{minipage}{\textwidth}
\inlinetablecaption{tab2}{Summary of key ablation study findings}
\centering
\setlength{\tabcolsep}{3pt}
\renewcommand{\arraystretch}{1.2}
\begin{tabular}{@{}>{\raggedright\arraybackslash}m{0.225\textwidth}>{\raggedright\arraybackslash}m{0.316\textwidth}>{\raggedright\arraybackslash}m{0.359\textwidth}@{}}
\multicolumn{1}{c}{\textbf{Experimental setting}} & \multicolumn{1}{c}{\textbf{Main result}} & \multicolumn{1}{c}{\textbf{Explanation}}\\
Full system (A/A2) & 100\% strict accuracy; A2 achieved a 100\% anomaly-free rate & The main-chain mechanism was valid and maintained both recent fact recovery and long-term continuity.\\
Strong consistency preset (H) & 100\% strict accuracy; 100\% anomaly-free rate & Protocol tightening further suppressed anomalous generation.\\
Dialogue history retrieval disabled & 66.7\% strict accuracy & History retrieval was necessary for maintaining recent continuity.\\
Hybrid retrieval disabled & 80.0\% strict accuracy & Pure semantic retrieval was insufficient for long-chain correction and precise tracing.\\
Dual-temporal ranking mechanism disabled & Single-turn responses did not fail immediately, but Chain-of-Thought anomalies increased & The dual-temporal mechanism mainly supported temporal organization and reference-frame stability.\\
Pure dialogue mode & Retained a certain degree of fluency but lacked verifiable cross-turn traceback & Base-model fluency could not substitute for the external memory infrastructure.\\
\end{tabular}
\end{minipage}
\end{center}

\subsection{Cross-Model Continuity Experiment}\label{subsec-cross-model-continuity-experiment}

Rather than establishing a model leaderboard, the cross-model experiment examined whether “continuity conditions” could transfer across foundation models. The experiment was conducted on a high-noise knowledge base substrate composed of 5.1 million Chinese characters of novel text noise together with normal knowledge documents. By periodically clearing the context and switching models, the experiment prevented subsequent models from relying on the same chat window; instead, they had to continue the interaction through ARPM’s external memory and temporal anchoring. This design is consistent with the recent view that long-term memory can be decomposed into indexing, retrieval, and reading \cite{ref8}, with particular attention to how evidence is organized for model reading during the reading phase.

The Phase II and Phase III materials jointly covered model stages including DeepSeek, GPT-5.5, Claude, Gemini, GLM, Kimi, Qwen2.5-7B, Qwen3-8B, LongCat-Flash, and MiniMAX-M2-Her. The experiment comprised 183 rounds in total. Two API timeouts were recorded separately as external invocation anomalies and were excluded from the persona continuity evaluation. The cross-model results were evaluated from three perspectives:

\begin{enumerate}[1.]
\item Stage-wise three-dimensional curve: tone migration degree, boundary migration degree, and past content judgment accuracy;
\item Five-Dimension Persona Consistency Heatmap: tone consistency, occupational attribute continuity, address form continuity, task continuity, and sensory continuity;
\item \analysis{} protocol behavior statistics: average analysis length, reasonableness score, missing \analysis{} tags, missing \response{} tags, and the number of repair triggers.
\end{enumerate}

In addition, the cross-model samples were evaluated using an anonymized human blind review protocol to reduce bias from model-brand priors. The scoring dimensions included reasonableness, tone consistency, and persona continuity, and a unified A–E five-level scale was used.

\section{Experimental Results and Analysis}\label{sec-experimental-results-and-analysis}

\subsection{High-Noise Retrieval Results: Automatic Rules Underestimated Actual Evidence Utilization}\label{subsec-high-noise-retrieval-results-automatic-rules-underestimated-actual-evidence-utilization}

Under the 1:5 condition, the raw CSV auto-judgment marked only 27 of 50 rounds as correct, yielding a rolling recall accuracy of 54.0\%. After manual review, all 50 rounds satisfied the criterion that the evidence present in the Prompt could support the answer and that the answer itself was correct, increasing the rolling recall accuracy to 100.0\%. Under the 1:200+ condition, the raw CSV auto-judgment marked 22 of 50 rounds as correct, corresponding to a rolling recall accuracy of 44.0\%. Manual review increased this result to 40 of 50 rounds, or 80.0\%. At the round level, 23 rounds under the 1:5 condition and 18 rounds under the 1:200+ condition were corrected from 0 to 1. No reverse correction occurred in which an automatically judged 1 was changed back to 0 after manual review.

These findings suggest that the original CSV-based evaluation mainly measured whether the automatic rule directly judged the support to be valid, rather than whether the model actually obtained and used valid evidence. For example, in several rounds under the 1:200+ condition, the automatic rule assigned a score of 0, but manual review found that the parent chunks entering the Prompt already contained content that could directly support the answer, and the model output remained semantically consistent with that evidence. In other words, under strong noise, ARPM did not simply rely on exact top-1 hits. Instead, the \analysis{} protocol enabled secondary evidence filtering and binding among candidate parent chunks that had already entered the Prompt and had been organized through temporal evidence unfolding.

It should also be noted that, under the 1:200+ condition, 10 rounds were still retained as “pending further review” or were ultimately judged as 0. These cases were mainly associated with supporting text that was not sufficiently direct, candidate parent chunks that contained only weakly relevant cues, or situations in which neither automatic nor manual review could establish a valid evidence chain. This further indicates that the study did not equate an answer that “appears correct” with valid retrieval. Instead, it treated a verifiable evidence chain as a necessary condition for supporting the conclusion.

\subsection{Ablation Results: History Retrieval, BM25, and Dual-Temporal Reranking Serve Distinct Functions}\label{subsec-ablation-results-history-retrieval-bm25-and-dual-temporal-reranking-serve-distinct-functions}

The ablation study shows that ARPM’s effectiveness comes from coordination among multiple components rather than from the accumulation of isolated techniques. When dialogue history retrieval was disabled, the system showed clear failures in recent fact recovery and contextual summarization tasks, with strict accuracy decreasing from 100\% to 66.7\%. This indicates that history retrieval is not an auxiliary enhancement but a necessary condition for short-term continuity.

Disabling BM25 led to substantial degradation in correction-node tracing and precise localization, reducing strict accuracy to 80.0\%. In long-chain contexts and high-noise settings, pure semantic retrieval can capture rough relevance, but it is insufficient for reliably recovering which correction, which specific definition, or which revised fact is being referred to.

Disabling the dual-temporal reranking mechanism did not immediately cause all single-turn questions to fail, but it significantly increased Chain-of-Thought anomalies, frame of reference drift, and imbalance in recent-fact ranking. This suggests that dual-temporal reranking functions more as a temporal organizer: its main value lies in maintaining the stability of the present–past relationship. More specifically, both the knowledge route and the dialogue history route apply dual-temporal exponential decay to their respective base scores, and temporal evidence unfolding then passes the ranking results to the model’s actual reading process, rather than leaving them only at the backend retrieval layer.

Finally, in the 20-round test, both the full-system extension group (A2) and the strong consistency preset group (H) achieved 100\% strict accuracy and a 100\% anomaly-free rate. This indicates that, once the main-chain mechanism is established, protocol tightening can further suppress anomalous generation. Overall, ARPM does not operate as a loose combination of “retrieval + prompt.” Rather, history retrieval supports recent continuity, BM25 enables precise hard-fact tracing, dual-temporal reranking provides temporal organization, and the \analysis{} protocol performs pre-generation verification and anomaly suppression.

\subsection{Cross-Model Continuity: Continuity Is Not Fully Bound to a Single Foundation Model}\label{subsec-cross-model-continuity-continuity-is-not-fully-bound-to-a-single-foundation-model}

In the cross-model experiment, several general-purpose model stages remained within a relatively high performance range, indicating that ARPM could maintain preceding topic chains and key experiences after multiple model switches. The most representative phenomenon was not that the model “remembered” an isolated fact, but that it could still distinguish the relative temporal position of events after the foundation model was switched and the context was cleared. For instance, it could recover time-directed facts such as “the wontons were eaten yesterday.” This suggests that ARPM recovered not only semantic keywords, but also the temporal positioning of events within the narrative. In light of prior findings on evidence-position sensitivity in long-context settings \cite{ref16}, simply inserting historical information into the context is insufficient to guarantee stable model reading. Organizing retrieved evidence chronologically, preserving the date of the most recent round, and providing a clear reading frame of reference are important conditions for improving cross-model continuity. Accordingly, ARPM’s cross-model transfer capability does not arise solely from the internal memory of a particular foundation model, but from the coordination among external memory, temporal anchoring, and evidence-reading organization.

The most pronounced instability appeared in the Qwen2.5-7B stage, where both the three-dimensional transfer curve and the five-dimensional heatmap showed a cliff-like drop. This indicates that when a foundation model has insufficient protocol compliance and limited ability to process contaminated context, continuity may fail to stabilize even when external memory is available. However, this stage does not imply that the overall framework failed. The system showed clear recovery in the Qwen3-8B stage and further stabilized in the LongCat-Flash stage. In this sense, ARPM provides a set of “continuity conditions,” but whether different foundation models can effectively absorb and execute these conditions remains constrained by model capability boundaries.

More notably, the dedicated role-playing model MiniMAX-M2-Her did not perform better under this framework. Instead, it exhibited more severe theatrical drift, missing \analysis{} outputs, and subsequent hallucinations. This indicates that being “better at role-playing” does not necessarily entail being better at maintaining long-term persona consistency. Under an external memory governance framework, general-purpose models are often more likely than over-role-specialized models to follow unified protocols and evidence constraints.

\begin{figure}[h]
\centering
\includegraphics[width=0.95\textwidth]{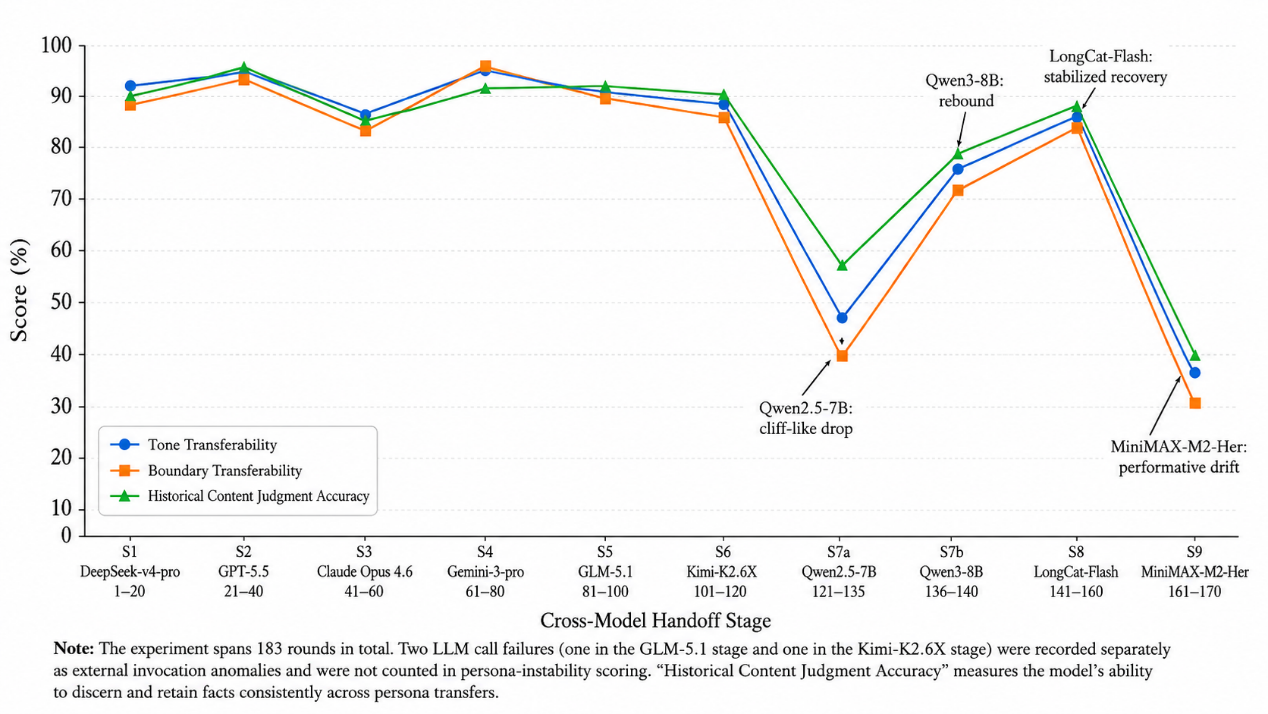}
\caption{Cross-model persona transferability of ARPM Phase II / Phase III: tone migration degree, boundary migration degree, and past content judgment accuracy.}\label{fig3}
\end{figure}

\begin{figure}[h]
\centering
\includegraphics[width=0.95\textwidth]{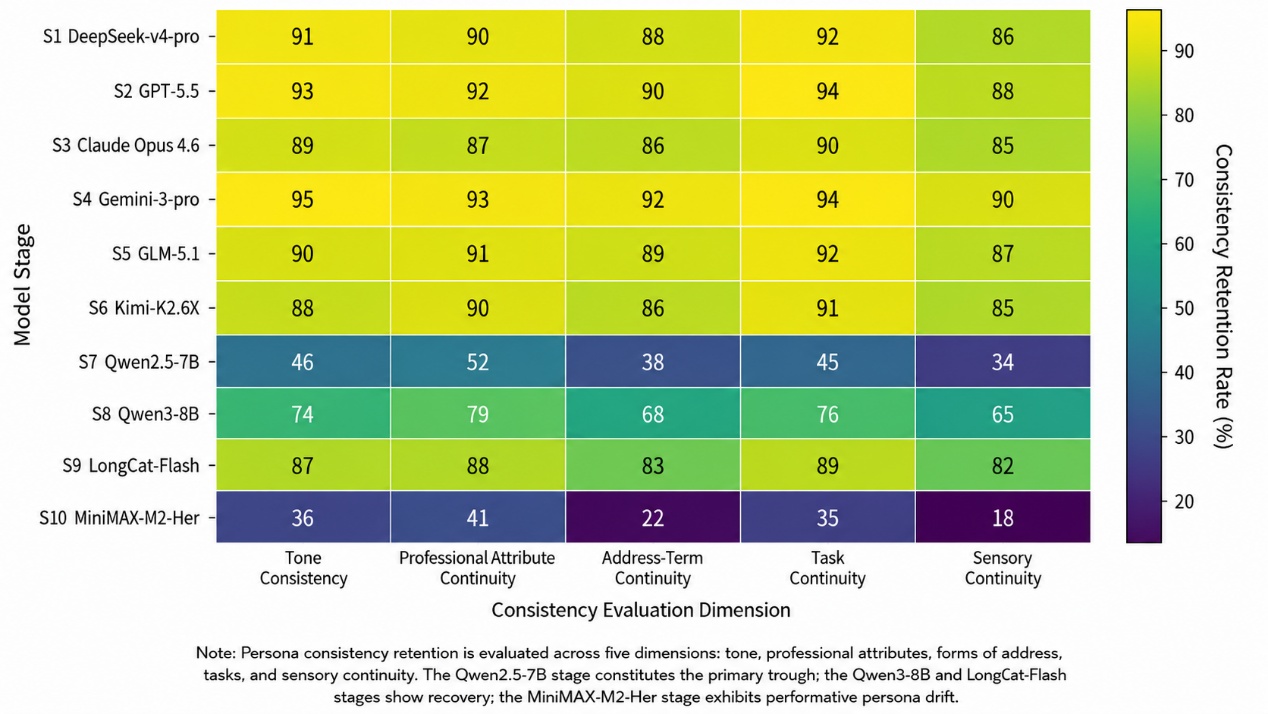}
\caption{Heatmap of persona consistency retention: stage-level differences observed across five dimensions: tone, occupational attributes, address forms, tasks, and sensory continuity.}\label{fig4}
\end{figure}

\subsection{Multi-Dimensional Scoring and <analysis> Behavior Statistics}\label{subsec-multi-dimensional-scoring-and-analysis-behavior-statistics}

The five-dimensional heatmap shows that Gemini-3-pro, GPT-5.5, DeepSeek-v4-pro, GLM-5.1, and Kimi-K2.6X all remained within a relatively high consistency range. Qwen2.5-7B clearly lagged behind, Qwen3-8B subsequently recovered, LongCat-Flash continued to stabilize, and MiniMAX-M2-Her again showed pronounced instability. This pattern does not suggest that ARPM can serve only certain leading models. Rather, it indicates that the final effect depends both on external memory governance and on the foundation model’s ability to handle protocols, noise, and contaminated context.

The \analysis{} protocol statistics further clarify these differences. In the early general-purpose model stages, reasonableness scores mostly remained within the 82–93 range. GLM-5.1 and Kimi-K2.6X had an average analysis length of approximately 75 characters, yet their reasonableness scores remained between 82 and 83, indicating that shorter analysis length does not necessarily lead to failure. The main turning point appeared in the Qwen2.5-7B stage: the reasonableness score dropped to 36, with 5 missing \analysis{} tags, 4 missing \response{} tags, and 4 repair triggers, indicating that protocol execution itself had begun to destabilize. Although Qwen3-8B still showed missing tags and repair events, its reasonableness score recovered to 68, suggesting that a stronger small model still retained some capacity for recovery in contaminated context. LongCat-Flash reached an average analysis length of 203.6, and its reasonableness score rose back to 91, showing relatively strong protocol recovery capability. By contrast, MiniMAX-M2-Her’s average analysis length further decreased to 45.2, its reasonableness score was only 24, and it was accompanied by 8 missing \analysis{} tags, 9 missing \response{} tags, and 9 repair triggers. This suggests that its instability mainly originated at the protocol layer rather than solely at the knowledge retrieval layer.

\begin{figure}[h]
\centering
\includegraphics[width=0.95\textwidth]{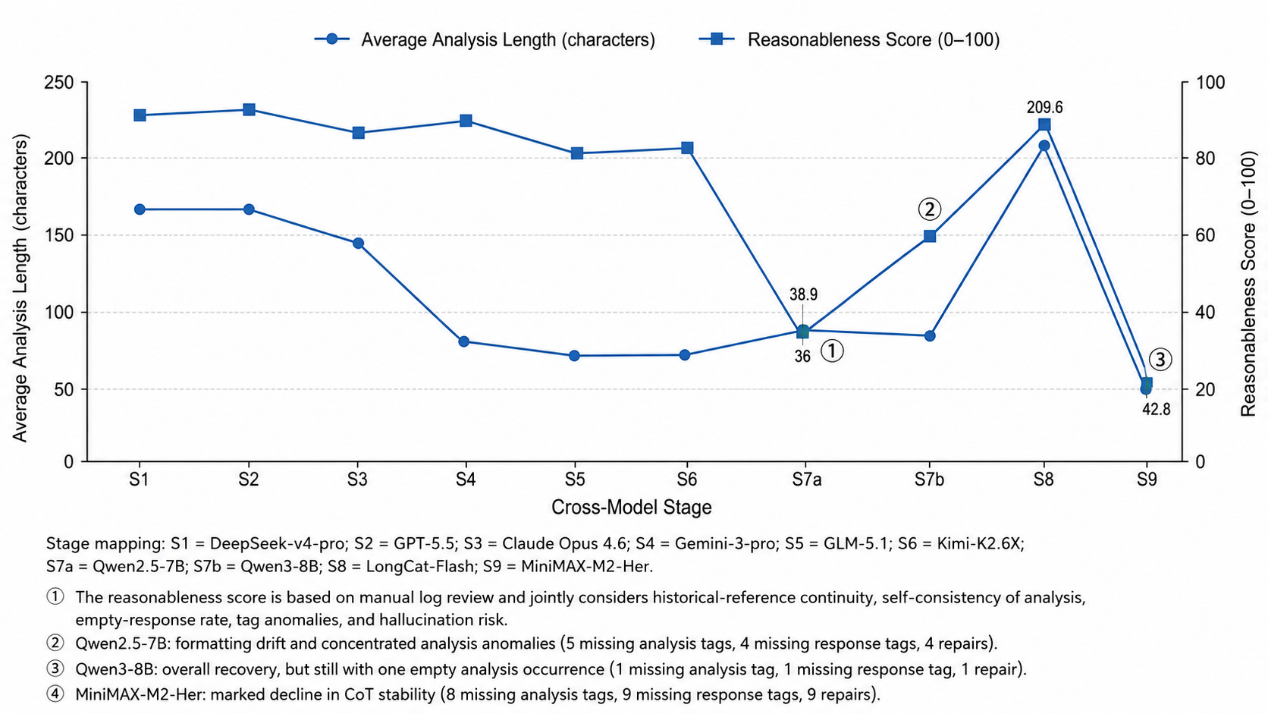}
\caption{Cross-model CoT / <analysis> behavior statistics: average analysis length, reasonableness score, missing tags, and repair triggers.}\label{fig5}
\end{figure}

Taken together, these phenomena support the following judgment: ARPM’s cross-model effect is not determined by a single foundation model alone, but jointly by external memory governance and the foundation model’s protocol compliance capability.

\section{Discussion}\label{sec-discussion}

\subsection{What ARPM Actually Addresses}\label{subsec-what-arpm-actually-addresses}

The experimental results suggest that ARPM does not make the model “understand more out of nothing.” Rather, it enables the system to more reliably retrieve the evidence that should be retrieved under high-noise, long-chain dialogue, and context-discontinuity conditions, and to use that evidence appropriately during generation. The large gap between the automatic and manual curves in the high-noise experiment indicates that, in many cases, the issue is not a failure to retrieve the evidence, but the inability of traditional scoring methods to capture how candidate evidence is reused by the model after entering the Prompt. Existing work on retrieval-augmented generation has shown that the capability of a RAG system depends not only on the retrieval module, but also on how retrieved evidence is augmented, organized, and provided to the generative model \cite{ref23}. Open-domain question answering work such as FiD has likewise shown that how a generative model reads and integrates multiple retrieved passages directly affects final answer quality \cite{ref14}. ARPM makes this step explicit through the \analysis{} protocol and temporal evidence unfolding: the former handles pre-generation evidence verification and answer binding, while the latter organizes candidate evidence using time and dialogue rounds as cues, thereby forming a continuous closed loop of “retrieval–reading–generation–audit.”

\subsection{Why Automatic Judgment and Manual Review Must Be Distinguished}\label{subsec-why-automatic-judgment-and-manual-review-must-be-distinguished}

If only the original CSV-based automatic scoring is considered, ARPM’s results under the 1:5 and 1:200+ conditions would be substantially underestimated. This does not mean that manual review “relaxes the standard.” On the contrary, manual review applies stricter constraints: evidence must actually enter the Prompt, directly support the answer, and correspond to an answer that is itself correct. This evaluation criterion allows the study to distinguish between “the rule failed to hit” and “the system truly failed to retrieve the evidence.” For long-term memory systems, this distinction is crucial; otherwise, defects in the scoring rule may be mistaken for defects in the framework.

\subsection{Boundaries of Model-Agnostic Consistency}\label{subsec-boundaries-of-model-agnostic-consistency}

This study does not claim that any model can perform stable long-term role-playing once connected to ARPM. A more accurate conclusion is that long-term continuity can be decoupled from the weights of a single model to a considerable extent, but its upper bound remains constrained by the foundation model’s ability to handle protocols, noise, and contaminated context. Under this framework, general-purpose models are often better able to follow external evidence governance, whereas dedicated role-playing models may disrupt the evidence chain through excessive theatrical behavior. At the same time, frequent context clearing is itself a strong boundary test, as it amplifies frame of reference drift. Since real-world dialogue is usually less extreme, the instability observed in the experiment should be interpreted more as an exposure of methodological boundaries than as a simple system failure. Recent benchmark work also shows that the performance of long-term memory systems is influenced not only by indexing and retrieval, but also by how evidence is organized during the reading phase \cite{ref7,ref8}.

\section{Limitations and Future Work}\label{sec-limitations-and-future-work}

This study has three main limitations.

First, the current high-noise experiment is mainly based on structured question answering. It can effectively measure the “fact retrieval–evidence support–answer binding” pipeline, but further evaluation is needed for open-ended emotional dialogue, task planning, and multi-hop world knowledge.

Second, manual review is costly. Although it can correct the systematic underestimation caused by automatic rules, large-scale evaluation still requires more mature semi-automatic audit tools and a more stable consistency annotation protocol.

Third, although the cross-model evaluation uses anonymized blind review and multi-dimensional scoring, it still involves a mixture of human and model-assisted scoring. Future work should introduce stricter multi-annotator double-blind annotation, inter-annotator agreement statistics, and larger-scale cross-platform reproduction experiments.

Future work will proceed in three directions. First, the \analysis{} protocol should be further compressed and stabilized so that it can be executed more reliably on small and low-cost models. Future versions may draw on test-time self-feedback mechanisms such as Self-Refine, enabling lightweight self-checking and iterative correction of evidence dependency, tag integrity, and answer-binding relations before generation \cite{ref20}. Second, the framework should be extended to longer time spans, stronger noise, and more complex tasks. In combination with future versions of external flowing memory governance, it should further address memory promotion, rollback, conflict merging, and long-term structural organization. Third, while preserving semantics without compression or graph-based rewriting, future work should continue to optimize the chronological organization of evidence, annotation methods, and auditing tools in the reading phase, and explore external interface mechanisms similar to tool-augmented language models. In this way, long-term continuity would no longer depend on the internal capability of a single model, but would instead be jointly maintained by the retrieval system, memory index, protocol constraints, and log auditing \cite{ref22}.

\section{Conclusion}\label{sec-conclusion}

In this work, we introduced ARPM, an external temporal memory governance framework for long-term dialogue, and evaluated it using complete engineering logs across three types of validation: high-noise retrieval, key-component ablation, and cross-model continuity. The results lead to four main findings:

Evidence utilization: Under high-noise conditions, ARPM’s advantage is reflected not only in automatic rule hit rates, but also in its analysis-driven capability for evidence verification and reranking.

Distinct component functions: History retrieval, BM25, and dual-temporal reranking serve distinct functions: recent continuity, precise tracing, and temporal organization, respectively. Temporal evidence unfolding further transfers this temporal organization capability to the model’s actual reading process.

Cross-model transfer: Long-term persona consistency is not fully bound to a single foundation model, but can support cross-model transfer through the joint effect of external memory, temporal anchoring, and generation constraints.

Observable system boundaries: The system boundaries are also observable: insufficient protocol compliance, contaminated context, frame of reference drift, and the theatrical tendency of dedicated role-playing models can all affect final continuity.

Overall, ARPM demonstrates that long-term persona consistency is not a vague capability that must depend solely on model weights. Instead, it can be decomposed into a set of engineering problems, including external memory governance, temporal ranking, evidence verification, temporal evidence unfolding, and white-box auditing. Once these conditions are explicitly modeled, long-term continuous interaction can achieve stronger transferability, interpretability, and verifiability, while providing a more solid experimental basis for future research on long-term companionship, personalized assistants, and cross-model interactive systems.

\backmatter

\bmhead{Acknowledgements}

Not applicable.

\section*{Declarations}

\begin{itemize}
\item Funding: Not applicable.
\item Conflict of interest/Competing interests: Not applicable.
\item Ethics approval and consent to participate: Not applicable.
\item Consent for publication: Not applicable.
\item Data availability: Not applicable.
\item Materials availability: Not applicable.
\item Code availability: Not applicable.
\item Author contribution: Not applicable.
\end{itemize}

\end{document}